\pdfoutput=1

\documentclass[11pt]{article}

\usepackage[final]{acl}

\usepackage{times}
\usepackage{latexsym}

\usepackage[T1]{fontenc}

\usepackage[utf8]{inputenc}

\usepackage{microtype}

\usepackage{inconsolata}

\usepackage{amsmath}
\usepackage{algorithm}
\usepackage{algorithmic}
\usepackage{enumitem}
\usepackage{multirow}
\usepackage{subcaption}
\usepackage{xspace}
\usepackage{booktabs}
\usepackage{graphicx}
\usepackage{listings}
\usepackage{xcolor}
\usepackage{courier}

\definecolor{codegreen}{rgb}{0,0.6,0}
\definecolor{codegray}{rgb}{0.5,0.5,0.5}
\definecolor{codepurple}{rgb}{0.58,0,0.82}
\definecolor{backcolour}{rgb}{0.95,0.95,0.92}

\lstdefinestyle{mystyle}{
    backgroundcolor=\color{backcolour},
    commentstyle=\color{codegreen},
    keywordstyle=\color{magenta},
    numberstyle=\tiny\color{codegray},
    stringstyle=\color{codepurple},
    basicstyle=\ttfamily\footnotesize,
    breakatwhitespace=false,
    breaklines=true,
    captionpos=b,
    keepspaces=true,
    numbers=left,
    numbersep=5pt,
    showspaces=false,
    showstringspaces=false,
    showtabs=false,
    tabsize=2
}
\newcommand{\prevtool}[0]{\textsf{{MEGAnno}}\xspace}
\newcommand{\tool}[0]{\textsf{{MEGAnno+}}\xspace}

\title{\tool : A Human-LLM Collaborative Annotation System}

\author{Hannah Kim, Kushan Mitra, Rafael Li Chen, Sajjadur Rahman, Dan Zhang \\
Megagon Labs \\ 
\texttt{\{hannah, kushan, rafael, sajjadur, dan\_z\}@megagon.ai}}

\begin{document}
\maketitle


\begin{abstract}
Large language models (LLMs) can label data faster and cheaper than humans for various NLP tasks.
Despite their prowess, LLMs may fall short in understanding of complex, sociocultural, or domain-specific context, potentially leading to incorrect annotations.
Therefore, we advocate a collaborative approach where humans and LLMs work together to produce reliable and high-quality labels.
We present \tool, a human-LLM collaborative annotation system that offers effective LLM agent and annotation management, convenient and robust LLM annotation, and exploratory verification of LLM labels by humans.
\footnote{
Demo \& video: \url{https://meganno.github.io}
}
\end{abstract}


\section{Introduction}
\label{sec:intro}

Data annotation has long been an essential step in training machine learning (ML) models. Accurate and abundant annotations significantly contribute to improved model performance. Despite the recent advancements of pre-trained Large Language Models (LLM), high-quality labeled data remains crucial in various use cases requiring retraining. For instance, distilled models are often deployed in scenarios where repeated usage of LLMs for inference can be too costly (e.g., API calls) or time-consuming (e.g., hosting on-premise). In specialized domains like medical and human resources, organizations often need customized models to meet heightened accuracy requirements and ensure the privacy of sensitive customer data. In addition to the training step, accurate labeled data is also necessary for evaluating and understanding of model performance.

Recent explorations~\cite{wang-etal-2021-want-reduce,ding-etal-2023-gpt} have showcased the potential of LLMs in automating the data annotation process. 
Unlike previous task-specific machine learning models, LLMs exhibit remarkable flexibility to handle any textual labeling task as long as suitable prompts are provided.
Besides, compared to traditional annotation relying solely on human labor, LLMs can usually generate labels faster and at a lower cost.
For example, hiring crowd workers for labeling may encounter problems such as delays, higher cost, difficulty in quality control~\cite{douglas2023data, sheehan2018crowdsourcing, litman2021reply, garcia2016challenges}.
Studies~\cite{Gilardi2023} show that LLMs can achieve near-human or even better-than-human accuracy in some tasks.
Furthermore, downstream models trained with LLM-generated labels may outperform directly using an LLM for inference~\cite{wang-etal-2021-want-reduce}.

Despite these advancements, it is essential to acknowledge that LLMs have limitations, necessitating human intervention in the data annotation process. 
One challenge is that the performance of LLMs varies extensively across different tasks, datasets, and labels~\cite{zhu2023can,ziems2023large}.
LLMs often struggle to comprehend subtle nuances or contexts in natural language, making involvement of humans with social and cultural understanding or domain expertise crucial.
Additionally, LLMs may produce biased labels due to potentially biased training data~\cite{Abid2021,sheng-etal-2021-societal}. In such cases, humans can recognize potential biases and make ethical judgements to correct them.

In this work, we present \tool, an annotation system facilitating human-LLM collaboration through efficient LLM annotation and selective human verification.
While LLM annotations are gaining interest rapidly, a comprehensive investigation on how to onboard LLMs as annotators within a human-in-the-loop framework in labeling tools has not been conducted yet.
For example, supporting LLM annotation requires not only user-friendly communications with LLMs, but also a unified backend capable of storing and managing LLM models, labels, and additional artifacts.
Efficient human verification calls for a flexible search and recommendation feature to steer human efforts towards problematic LLM labels, along with a mechanism for humans to review and rectify LLM labels.
Throughout the paper, we explain how we achieve this in our system, showcase a use case, and discuss our findings.

We summarize our contributions as below:
\vspace{-0.3em}
\begin{itemize}
  \setlength{\itemsep}{-.3\baselineskip}
  \item A human-LLM collaborative annotation system that offers 1) effective management of LLM agents, annotations, and artifacts, 2) convenient and robust interfacing with LLMs to obtain labels, and 3) selective, exploratory verification of LLM labels by humans.
  \item A use case demonstrating the effectiveness of our system.
  \item Practical considerations and discussion on adopting LLMs as annotators.
\end{itemize}


\section{Related Work}
\label{sec:relwork}

\paragraph{LLMs as annotators}
There is growing interest in utilizing LLMs as general-purpose annotators for natural language tasks~\cite{kuzman2023chatgpt,zhu2023can,ziems2023large}.
\citet{wang-etal-2021-want-reduce} find that GPT-3 can reduce labeling cost by up to 96\% for classification and generation tasks.
Similarly, \citet{ding-etal-2023-gpt} evaluate GPT-3 for labeling and augmenting data in classification and token-level tasks.
Other studies show that for some classification tasks, LLMs can even outperform crowdsourced annotators~\cite{Gilardi2023,he2023annollm,tornberg2023chatgpt4}.

\paragraph{Verification of LLM responses}
To detect and correct erroneous responses from LLMs, approaches to rank or filter LLM outputs have been explored. 
The most common method is using model logits to measure model uncertainty~\cite{wang-etal-2021-want-reduce}.
More recently, \citet{lapras} propose training a verifier model using various signals from LLMs' input, labels, and explanations.
Alternative methods include asking LLMs to verbalize confidence scores~\cite{lin2022teaching} and calculating consistency over prompt perturbations~\cite{wang2023selfconsistency,xiong2023llms}.
Other line of works investigate self-verification, i.e., LLMs give feedback on their own outputs and use them to refine themselves~\cite{madaan2023selfrefine,cheng2023blackbox}.
In our system, we focus on human verification of LLM-generated labels and leave model verification and self-verification as future work.

\paragraph{Annotation tools with AI/ML assistance}
Machine learning models have proven effective in assisting humans in various steps of the training data collection pipeline. Annotation tools and frameworks such as Prodigy~\cite{montani2018prodigy}, HumanLoop~\cite{humanLoop}, Label Studio~\cite{LabelStudio}, and Label Sleuth~\cite{shnarch-etal-2022-label} all aim to enhance the subset selection step with active learning approaches. ML models are also naturally used to make predictions, serving as pre-labels. 
For instance, INCEpTION~\cite{klie-etal-2018-inception} provides annotation suggestions generated by ML models.
HumanLoop~\cite{humanLoop} and Autolabel~\cite{autolabel} support the annotation or augmentation of datasets using either commercial or open-source LLMs. 
In this work, we go beyond using LLMs to assist annotation for human annotators or to replace human annotators.
Rather, \tool advocates for a \textit{collaboration} between humans and LLMs with our dedicated system design and annotation-verification workflows.


\section{Design Considerations}
\label{sec:designgoals}

Let us start with a motivating example of Moana, a Data Scientist working at a popular newspaper.
Moana is tasked with training a model to analyze the degree of agreement between user comments and political opinion pieces --- e.g., whether the comments entail the opinion.
Moana opts for LLM annotation, but she encounters various challenges in the process.
Firstly, without any guidance for prompting, she resorts to trial-and-error to eventually identify a suitable prompt for the task. Even so, she must perform additional validations to ensure that the annotated labels are within the space of pre-defined labels.
Moreover, the API calls to the LLM can be unreliable, throwing exceptions such as timing out and rate limit violations, requiring her to handle such errors manually. 
Next, Moana lacks the confidence to train a downstream model without verifying the LLM annotations. However, without any assistance in reviewing potential annotation candidates for verification, she has to go through all the annotations, which can be time-consuming.
Finally, she has to manually save used model configurations to reuse the model for additional datasets.

From Moana's example, we summarize our design requirements for a human-LLM collaborative annotation system as follows:

\begin{figure*}[t]
  \includegraphics[width=\textwidth]{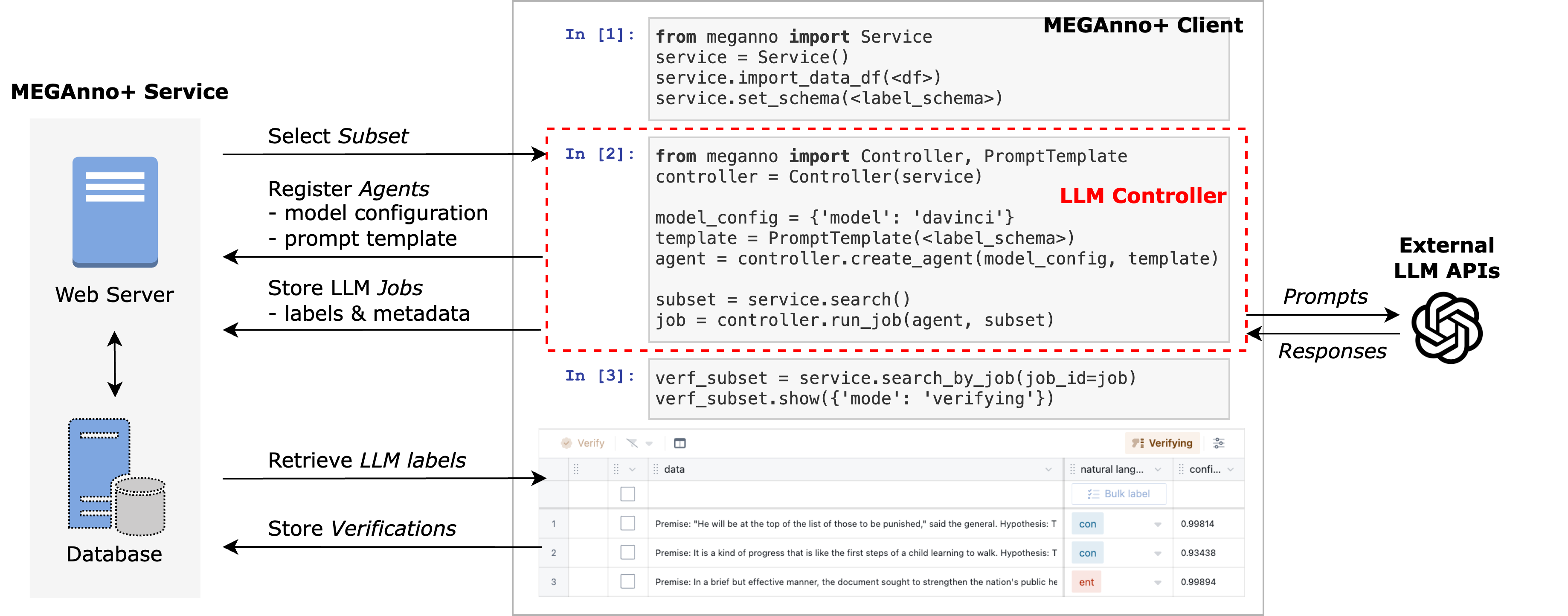}
  \caption{\tool system architecture and LLM-integrated workflow. With \tool client, users can interact with the back-end service that consists of web and database servers through programmatic interfaces and UI widgets. The middle notebook shows our workflow where cell [2] is LLM annotation and cell [3] is human verification.}
  \vspace{-.5em}
  \label{fig:architecture}
\end{figure*}

\begin{enumerate}[topsep=.3em, noitemsep]
  \item LLM annotation
  \begin{enumerate}
    \item \textbf{[Convenient]} Annotation workflow  including pre-processing, API calling, and post-processing is automated. \label{convenient}
    \item \textbf{[Customizable]} Flexibly modify model configuration and prompt templates. \label{customizable}
    \item \textbf{[Robust]} Resolvable errors are handled by the system. \label{robust}
    \item \textbf{[Reusable]} Store used LLM models and prompt templates for reuse. \label{reusable}
    \item \textbf{[Metadata]} LLM artifacts are captured and stored as annotation metadata. \label{meta}
  \end{enumerate}
  \item Human verification
  \begin{enumerate}
    \item \textbf{[Selective]} Select verification candidates by search query or recommendation. \label{selective}
    \item \textbf{[Exploratory]} Filter, sort, and search by labels and available metadata programmatically and in a UI. \label{exploratory}
    \end{enumerate}
\end{enumerate}

To satisfy these design requirements, we implement our system as an extension to \prevtool~\cite{zhang-etal-2022-meganno}, an in-notebook exploratory annotation tool.
Its flexible search and intelligent recommendations enable efficient allocation of human and LLM resources toward crucial data points (R~\ref{selective},\ref{exploratory}).
Additionally, \prevtool provides a cohesive backend for the storage of data, annotations, and auxiliary information (R~\ref{reusable},\ref{meta}). 


\section{System}
\label{sec:system}

\subsection{System Overview}
\label{sec:overview}

\tool is designed to provide a convenient and robust workflow for users to utilize LLMs in text annotation.
To use our tool, users operate within their Jupyter notebook~\cite{Kluyver2016jupyter} with the \tool client installed.

Our human-LLM collaborative workflow (Fig.~\ref{fig:architecture}) starts with LLM annotation.
This step involves compiling a subset and using an LLM to annotate it by interacting with the programmatic LLM controller. 
The LLM controller takes care of 1) agent registration and management (e.g., model selection and validation) and 2) running annotation jobs (e.g., input data pre-processing, initiating LLM calls, post-processing and storing responses), satisfying R~\ref{convenient},\ref{robust}. 
Once LLM annotation is completed, users can verify LLM labels.
Users can select a subset of LLM labels to verify by search queries (R~\ref{selective}), and inspect and correct them in a verification widget in the same notebook (R~\ref{exploratory}).

\paragraph{Data Model}
\tool extends \prevtool's data model where data \texttt{Record}, \texttt{Label}, \texttt{Annotation}, \texttt{Metadata} (e.g., text embedding or confidence score) are persisted in the service database along with the task \texttt{Schema}.\footnote{\tool only supports full LLM-integrated workflows for record-level tasks.} Annotations are organized around \texttt{Subsets}, which are slices of the data created from user-defined searches or recommendations.
To effectively integrate LLM into the workflow, we introduce new concepts: \texttt{Agent}, \texttt{Job}, and \texttt{Verification}.
An \texttt{Agent} is defined by the configuration of the LLM (e.g., model's name, version, and hyper-parameters) and a prompt template.
When an agent is employed to annotate a selected data subset, the execution is referred to as a \texttt{Job} (see Section~\ref{sec:job}). \texttt{Verification} captures annotations from human users that confirm or update LLM labels (see Section~\ref{sec:verification}).

\subsection{Agents: Model and Prompt Management}
\label{sec:agent}

Since variation in either model configuration or prompt may result in a variable output from an LLM, we define an annotation \texttt{Agent} to be a combination of a user-selected configuration and prompt template.
Used agents are stored in our database\footnote{Note that for an agent, we store its prompt template (a rule to build prompt text), not prompts (generated prompts for a set of data records) to save storage.} and can be queried based on model configuration.
This allows users to reuse agents and even compare the performance of different LLMs on a particular dataset (R~\ref{reusable}).

\paragraph{Model configuration}
\tool enables 
users to choose an LLM from a list of available models, configure model parameters, and also provide a validation mechanism to ensure the selected model and parameters conform to the LLM API definition and limitations.
While \tool is designed to support any open-source LLM or commercial LLM APIs, in this work, we only demonstrate OpenAI Completion models for clarity and brevity.

\begin{figure}[t]
  \includegraphics[width=\columnwidth]{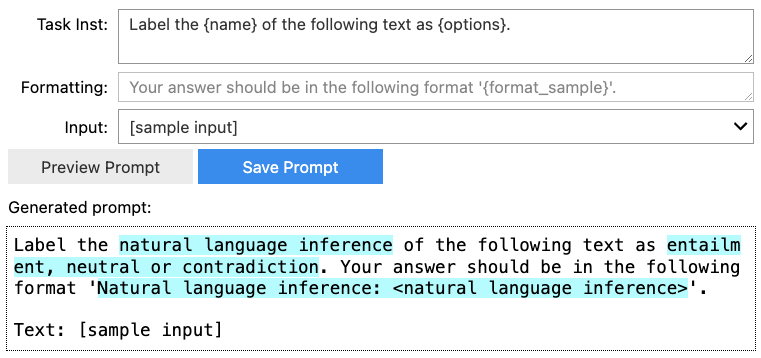}
  \caption{UI for customizing a prompt template and previewing generated prompts. 
  Prompt is generated based on the name and options of label schema.
  }
  \label{fig:prompt}
\end{figure}

\paragraph{Prompt template}
\label{sec:prompt}
To utilize LLMs as annotators, an input record has to be transformed into a prompt text.
With \tool, prompts can be automatically generated based on a labeling schema and a prompt template for users' convenience.
We offer a default template that contains annotation instruction, output formatting instruction, and input slot, which can be edited programmatically.
We also provide a UI widget to interactively customize the prompt template and preview the generated prompts for selected data samples (Fig.~\ref{fig:prompt}, R~\ref{customizable}).

\subsection{LLM Annotation}
\label{sec:controller}

Unlike human annotation, LLM annotation goes through a multi-step process to collect labels from input data records.
We execute this process as an annotation \texttt{Job} using the LLM controller.

\subsubsection{Initiating LLM Jobs}
\label{sec:job}

To start a job, users need to select a data subset to annotate and an agent, i.e., an LLM model and a prompt template.
Users can utilize \prevtool's sophisticated subset selection techniques, including filtering by keywords or regular expressions, or receiving suggestions of similar data records.
One can create a new agent or reuse one of previously registered agents.
By reusing subsets and agents for new jobs, users can easily compare annotation performance between different models or for different data slices.

\subsubsection{Pre-processing}
The first step within a job is pre-processing.
Using the prompt template of a selected agent, a data subset is converted into a list of prompts.
All prompts are validated (e.g., within max token limit) before calling LLM APIs.

\subsubsection{LLM API Calls: Error Handling}
\tool handles the calls to the external LLM APIs to facilitate a smooth, robust, and fault-tolerant experience for users, without having to worry about making any explicit API calls or handling error cases themselves. 
In order to ensure a fault-tolerant procedure, errors encountered during API calls are handled in two ways: \textit{handle within our system} or \textit{delegate to users}.
We handle known LLM API errors that can be solved by user-side intervention. 
This would be in cases such as a \texttt{Timeout} or \texttt{RateLimitError} in OpenAI models, or other similar errors which require the user themselves to call to the LLM API again. On encountering such errors, \tool retries the call to the LLM API itself. 
Delegated errors are the ones that require interventions by external service providers and are beyond our scope.
For instance, errors such as \texttt{APIConnectionError} in OpenAI models occur because of an issue with the LLM API server itself and requires intervention from OpenAI.
In this case, \tool simply notifies the user and relays the error message.

\subsubsection{Post-processing LLM Responses and Storing Labels and Metadata}

\begin{figure}[t]
  \centering
  \includegraphics[width=\columnwidth]{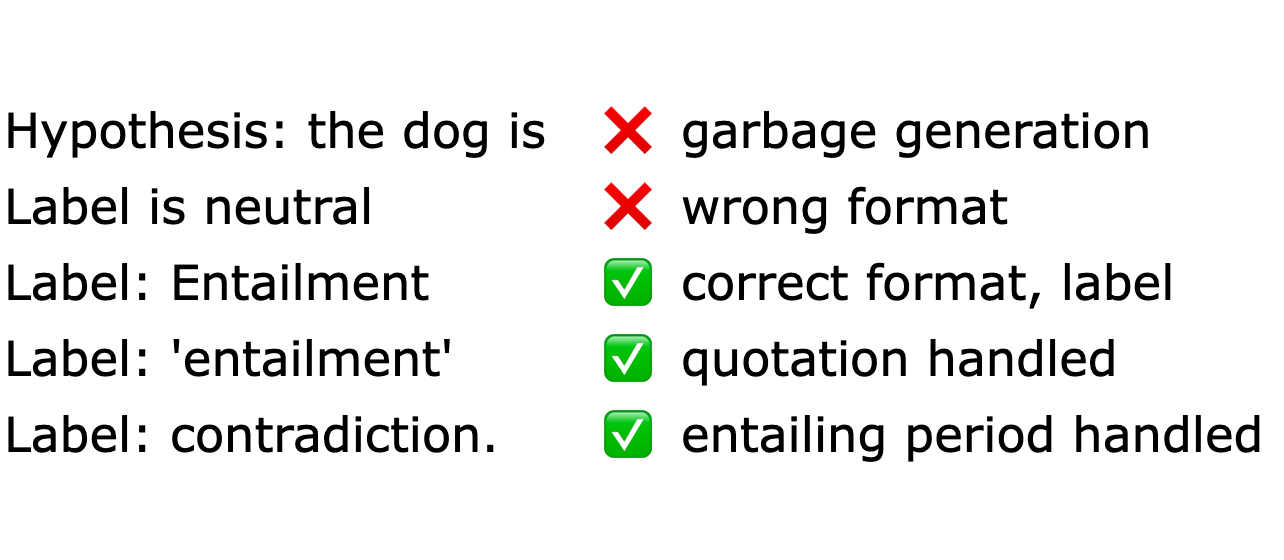}
  \vspace{-2em}
  \caption{Example LLM responses and extraction results. Minor violations are processed as valid labels.}
  \label{fig:extraction}
\end{figure}

\paragraph{Label extraction}
LLM outputs are typically unstructured (i.e., free-text) and can be noisy and unusable for downstream applications, even when prompted to adhere to a specific format. 
This necessitates careful post-processing of LLM-generated content, converting them into valid labels (Fig.~\ref{fig:extraction}).
\tool conducts an automated post-processing step on LLM responses, handling errors 
in cases of syntax or formatting violations (i.e., not adhering to the format specified in prompt instructions). Additionally, our tool checks for semantic violations, ensuring that the generated label is valid within the existing schema for the task.

\paragraph{Metadata extraction}
\tool can collect model artifacts and store them as label metadata (R~\ref{meta}).
Examples include model logits, costs associated with inference, used random seed, and so on.
They can be useful for further analyses on LLM annotation and human verification.
By default, our system only stores token logits to estimate the used LLM's confidence for generated labels.
Calculated confidence scores serve as additional signals for decision-making in the human verification step.

\paragraph{Storing in database}
Following the post-processing step, extracted valid labels and metadata are sent to the backend service for persistence in the database. Invalid labels are not stored in the database to prevent label contamination, but frequent invalid ones are still shown to the user to guide the next iteration (e.g., update labeling schema, improve instruction in prompts).

\begin{figure}[t]
  \includegraphics[width=\columnwidth]{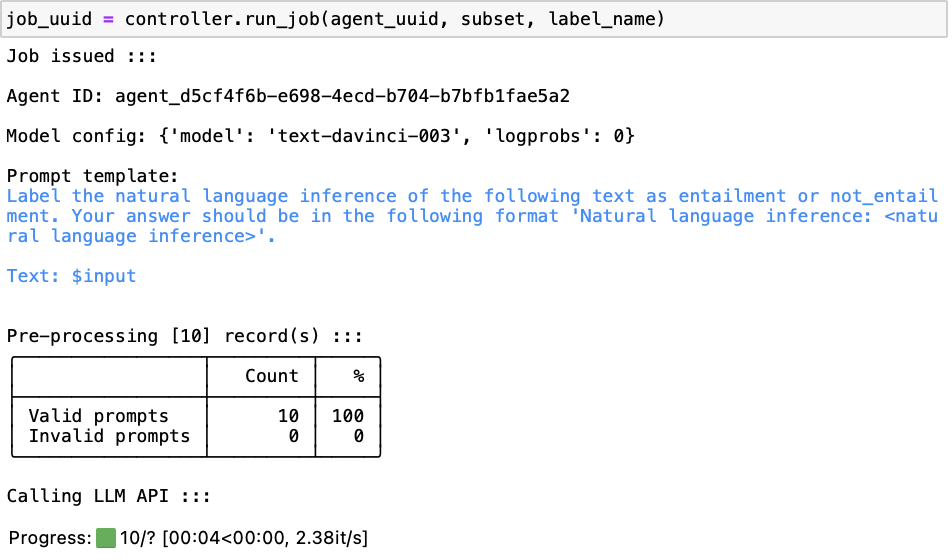}
  \includegraphics[width=.65\columnwidth]{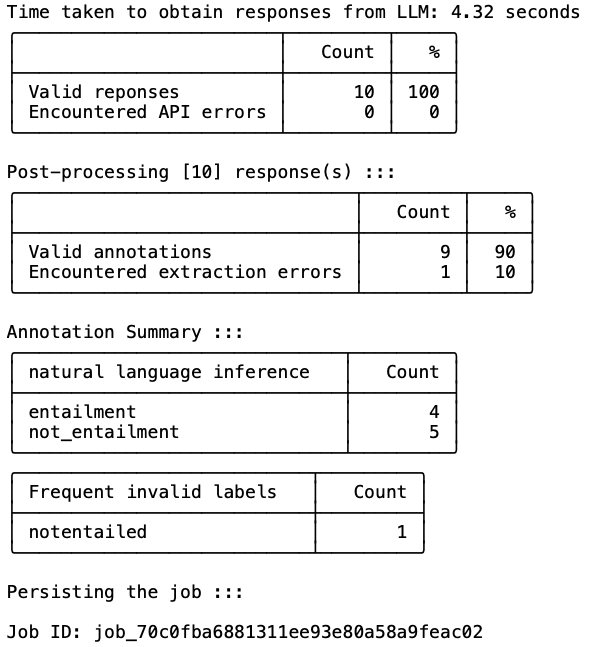}
  \caption{Annotation progress and summary.}
  \label{fig:summary}
\end{figure}

\subsubsection{Monitoring Annotation Jobs}
\label{sec:monitor}

When running a job, we display the progress and statistics of each step of the job for monitoring (Fig.~\ref{fig:summary}).
These include 1) agent details such as the selected model and prompt template, 2) input summary such as sample prompts generated using the template along with how many prompts are valid or invalid, 3) API call progress such as the time taken to retrieve responses from the API calls, and 4) output summary such as the numbers of valid and invalid responses from API and label distribution of valid responses.

\begin{figure}[t]
  \includegraphics[width=\columnwidth]{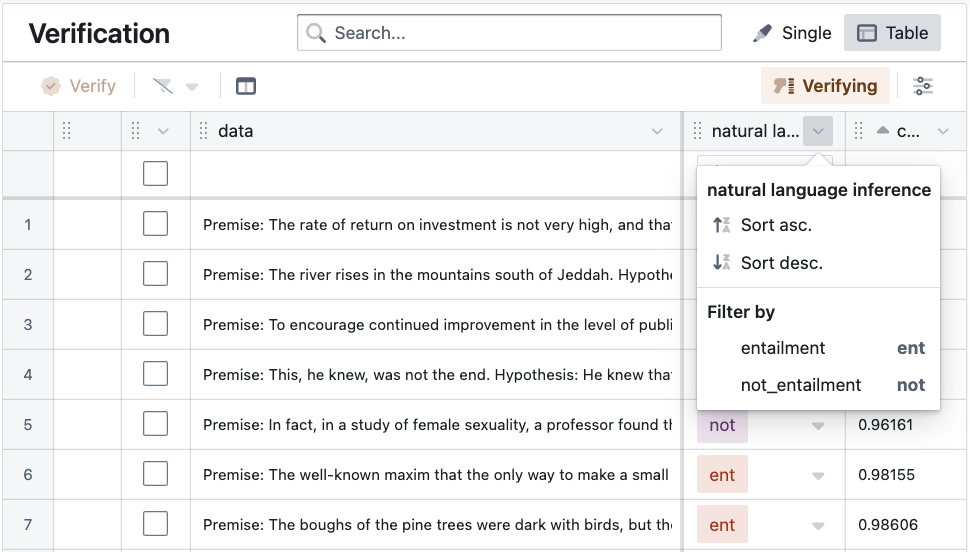}
  \caption{The table view in verification UI. Users can explore LLM annotations via filtering by labels, sorting by confidence scores, or keyword search on text input.}
  \label{fig:verification}
\end{figure}

\subsection{Verification}
\label{sec:verification}

LLM labels can be unreliable, requiring human verification to ensure the quality of the collected labeled data.
\paragraph {In-notebook verification widget}
\tool provides a verification widget to complete the LLM annotation workflow in the same notebook. Leveraging \tool's robust and customizable search functionality, users can retrieve a subset of LLM labels based on keywords, regular expressions, assigned labels, or metadata. Then utilizing the verification widget (as illustrated in Fig.~\ref{fig:verification}), users can explore the selected subset and decide whether to confirm or correct their LLM-generated labels. The verification UI includes both a table view for exploratory and batch verification, as well as a single view. 

\paragraph{Verification priority}
Human verification, while less expensive than direct annotation, can still be time- and cost-consuming. Therefore, it is crucial to prioritize and direct human efforts toward more ``suspicious'' outputs from LLMs. Our widget facilitates this process by presenting metadata, such as model confidence or token logit scores, in a separate column. Users can freely sort and filter rows based on labels or metadata, enabling them to prioritize or focus on labels with low confidence.

\paragraph{Query and export verified labels}
\tool offers flexible query interfaces, allowing users to search for verification by LLM agents (i.e., model and prompt config), as well as jobs. 
Both the original LLM-generated labels and any potential human corrections are stored in the database, enabling users to filter and retrieve labels ``confirmed'' or ``corrected'' by human verifiers. These features establish a foundation for easy in-notebook model and prompt comparison. Ultimately, the query results serve as a view of the labeling project, ready to be exported to downstream applications.


\section{Use Case: Natural Language Inference}
\label{sec:usecase}

Moana, the aforementioned data scientist who needs to collect training data quickly, decides to use \tool to leverage LLM-powered annotation.
First, she imports her unlabeled data and sets the labeling schema as entailment or not entailment.
She selects a GPT-3 davinci model with the default parameters and prompt template.
To test this setting, she runs the model on 10 samples.

\begin{lstlisting}[language=Python]
c = Controller(<service>, <auth>)
model_config = {'model': 'davinci'}
template = PromptTemplate(label_schema)
agent = c.create_agent(model_config, template)
subset = <service>.search(limit=10)
job = c.run_job(agent, subset)
\end{lstlisting}

\noindent
After the job is finished, the annotation summary (Fig.~\ref{fig:summary}) shows that all samples are successfully annotated by GPT-3 and 40\% are entailment.
Also, one response is annotated with `notentailed', exemplifying the instability of LLMs even with clear instructions.
With \tool's table view widget, she examines data and labels (Fig.~\ref{fig:verification}).
She realizes that some of the records labeled as `not entailment' are contradictory whereas the rest are neutral.
She updates the labeling schema to contain entailment, neutral, and contradiction.
Next, she wonders if changing the model's temperature would improve the accuracy of annotation.
She creates another agent, GPT-3 with temperature with zero and re-runs annotation on the same subset.

\begin{lstlisting}[language=Python]
model_config2 = {'model': 'davinci', 'temperature': 0}
agent2 = c.create_agent(model_config2, template)
job2 = c.run_job(agent2, subset)
\end{lstlisting}

\noindent
She exports the annotations from both jobs and compares them.
She concludes that the second model is good enough for her project.
She imports her entire data and uses the agent to label them.
Since the size of the data is huge, she has to wait till the annotations are done.
Fortunately, with \tool, she can track the progress in the output cell while the job is running.
To review the annotations, she sorts the annotations in an ascending order of confidence and manually verifies low confidence ($<95\%$) annotations.


\section{Discussion}
\label{sec:discussion}

\paragraph{How to design an annotation task?}
Based on our experience, we find that designing an annotation task and a prompt similar to more widely used and standardized NLP tasks is beneficial.
For example, framing Moana's problem as a natural language inference task is more effective than framing it as a binary classification of agreement and disagreement.
Also, the selection of label options may work better if it is similar to common options for given tasks, such as [positive, neutral, negative] > [super positive, positive, ..., negative] for sentiment classification.
Lastly, it is recommended that the format of a prompt be similar to the one used in training as some LLMs
have different prompt format than the others.
We plan to conduct more systematic test to discover reasonable default prompts for different models.

\paragraph{Are LLMs consistent and reliable annotators?}
We expect human annotators to maintain a consistent mental model.
In other words, when humans are presented with the same question rephrased, we anticipate consistent answers.
However, LLMs are known to be sensitive to semantic-preserving perturbations in prompts.
For instance, changes in prompt design, the selection and order of demonstrations, and the order of answer options can result in different outputs~\cite{poisoning:icml21,pezeshkpour2023large}.
Moreover, commercial LLMs can undergo real-time fine-tuning, meaning that prompting with the same setup today may yield different results than prompting yesterday~\cite{chen2023chatgpts}.
Therefore, LLM annotators and human annotators should not be treated the same, and annotation tools should carefully design their data models and workflows to accommodate both types of annotators.

\paragraph{Limitations}
Our system has several limitations.
Our post-processing mechanism may not be robust to cover all tasks and prompts entered by the user. 
Furthermore, \tool's ability to capture metadata is contingent on the LLM model used. For example, \texttt{GPT-4} models do not yet provide any form of token logprobs or other metadata which can be captured. 


\section{Conclusion}
\label{sec:conclusion}

\tool is a text annotation system for human-LLM collaborative data labeling.
With our LLM annotation $\rightarrow$ Human verification workflow, reliable and high-quality labels can be collected efficiently.
Our tool supports robust LLM annotation, selective human verification, and effective management of LLMs, labels, and metadata.

As future work, we are currently working on adding more LLM agents (e.g., open-source LLMs), supporting customized extraction of metadata (e.g., custom uncertainty metric), and improving prompt template UI for data-aware in-context learning.
Additionally, we plan to incorporate diverse annotation workflows such as Multi-agent LLM annotation $\rightarrow$ LLM label aggregation $\rightarrow$ Human verification; and LLM augmentation $\rightarrow$ Human verification.


\section*{Ethics Statement}
\label{sec:ethics}

First, labels generated by LLMs can exhibit bias or inaccuracy.
These models are pre-trained on vast amount of data, which are typically not accessible to the public.
Biases present in the training data can be transferred to LLM labels.
Also, if the training data lacks relevant or up-to-date knowledge, the model may produce incorrect annotation.
Since we cannot access models' inner workings or their training data, it is difficult to identify and understand how and why LLMs make biased or inaccurate labeling decisions.
Second, the use of commercial LLMs for labeling data containing sensitive information or intellectual property may pose risks.
Data shared with commercial LLMs, such as ChatGPT, may be collected and utilized for retraining these models.
To prevent potential data leakage and mitigate associated legal consequences, it is advisable to either mask any confidential information or only use in-house LLMs.


\section*{Acknowledgements}
We would like to thank Pouya Pezeshkpour and Estevam Hruschka for their helpful comments.

\bibliography{anthology_part1,anthology_part2,custom}

\appendix

\end{document}